\documentclass[journal]{IEEEtran}

\usepackage{graphicx}
\usepackage{amsmath}
\usepackage{subfigure}
\usepackage{tabularx}
\usepackage{cite}

\graphicspath{{Graphics/}}

\begin{document}

\title{Passivity Guaranteed Stiffness Control with Multiple Frequency Band Specifications for
	a Cable-Driven Series Elastic Actuator}

\author{Ningbo Yu*,
        Wulin Zou,
        and Yubo Sun
\thanks{This paper has been accepted and published in the journal \textbf{Mechanical Systems and Signal Processing}. Manuscript received April 21, 2018; revised July 3, 2018; accepted August 3, 2018.}
\thanks{Citation: Ningbo Yu, Wulin Zou, Yubo Sun, ``Passivity guaranteed stiffness control with multiple frequency band specifications for a cable-driven series elastic actuator'', \textit{Mechanical Systems and Signal Processing}, Vol. 117, pp. 709--722, 2019.}
\thanks{The authors are with the Institute of Robotics and Automatic Information Systems, Nankai University, and Tianjin Key Laboratory of Intelligent Robotics, Nankai University, Haihe Education Park, Tianjin 300350, China. Corresponding author: Ningbo Yu, E-mail: nyu@nankai.edu.cn.}
}

\maketitle

\begin{abstract}
Impedance control and specifically stiffness control are widely applied for physical human-robot interaction. 
The series elastic actuator (SEA) provides inherent compliance, safety and further benefits. 
This paper aims to improve the stiffness control performance of a cable-driven SEA. 
Existing impedance controllers were designed within the full frequency domain, though human-robot interaction commonly falls in the low frequency range. 
We enhance the stiffness rendering performance under formulated constraints of passivity, actuator limitation, disturbance attenuation, noise rejection at their specific frequency ranges. 
Firstly, we reformulate this multiple frequency-band optimization problem into the $H_\infty$ synthesis framework. 
Then, the performance goals are quantitatively characterized by respective restricted frequency-domain specifications as norm bounds.
Further, a structured controller is directly synthesized to satisfy all the competing performance requirements. 
Both simulation and experimental results showed that the produced controller enabled good interaction performance for each desired stiffness varying from 0 to 1 times of the physical spring constant. Compared with the passivity-based PID method, the proposed $H_\infty$ synthesis method achieved more accurate and robust stiffness control performance with guaranteed passivity.
\end{abstract}


\section{Introduction}

Physical human-robot interaction (HRI) is of fundamental importance for robotic research and has been greatly advanced over the last two decades~\cite{Haddadin2016,Tsarouchi2016}. To improve interaction safety and obtain inherent compliance during physical HRI, the series elastic actuation (SEA) structure in which an elastic component is intentionally placed between the motor and load was proposed~\cite{Pratt1995} and attracted continuous research efforts. The SEA provides a number of advantages over stiff actuation, including greater shock tolerance, more stable force output, lower reflected inertia, energy storage capacity and safety~\cite{Pratt1995}. 

Various SEAs have been developed and applied for physical human-robot interaction, such as the Bowden-cable-based SEA for the lower extremity powered exoskeleton (LOPES)~\cite{Vallery2008}, the MR-compatible SEA for wrist sensorimotor study~\cite{Sergi2015}, the SEA for a monopod hopping robot~\cite{WangMeng2016IROS}, the compact SEA for upper and lower limb rehabilitation~\cite{Li2017IJRR}, the Bowden-cable SEA for hand finger exoskeleton~\cite{Agarwal2015IJRR,Yun2016IROS,Agarwal2017JMR}, etc. The cable-driven SEA allows to detach the actuation motor from the robot frame, enables power transmission to remote place, and brings conveniences and flexibilities into system construction and control for applications to physical human-robot interaction~\cite{Veneman2006,Senturk2016,Zou2017JAS}.

The robot is required to have the capability of varying its behavior from being stiff to compliant and even transparent for different interaction tasks. Impedance, defined by the dynamic relationship between the robot's output torque and motion, well characterizes the stiffness/compliance of the interaction. Impedance control that was proposed by Hogan in~\cite{Hogan1984} has been a fundamental approach to shape the given system's behavior to match a predefined impedance model, and has been widely studied and applied in robotics~\cite{Tsuji2005,Ferraguti2013ICRA,HeWei2016,Calanca2016Tmech}. 

In this work, we aim to shape the impedance that the robot exhibits to the human, or equivalently, the impedance that the human perceives when interacting with the robot. In this case, the motion in the designed impedance model is the active motion that the human applies to the robot, and the torque/force is the interactive torque/force between the human and robot.

Conventional impedance control approaches design the controller in a cascaded manner and lots of such control strategies with a PID-based inner force/torque loop have been employed for SEA~\cite{Vallery2008,Sergi2015,Senturk2016,Jardim2014,Tagliamonte2014,Calanca2017}. Disturbance observer (DOB) based torque or impedance control strategies have also been proposed to improve SEA control accuracy and robustness~\cite{Paine2015JFR,YuHaoYong2015,Mehling2015IROS,Roozing2016IROS,Oh2017}. Recently, adaptive torque or impedance controllers have been developed to guarantee predictable performance despite uncertainties or disturbance in the SEA or human side~\cite{Calanca2014Robotica,Losey2016,Pan2017TCST,Kaya2017TIMC,Li2017TRO,Calanca2017RAS}.

In~\cite{Mehling2014,Santos2015,Yu2017AA}, the SEA impedance control structure was transformed into the $H_\infty$ control framework. The impedance controller can be synthesized directly with minimizing the impedance rendering error. 
With a full comprehension of the practical system, the performance requirements and physical constraints of the system can be transformed to corresponding quantified norm bounds to each signal of interest~\cite{ZhouKemin1998}.

Current SEA's impedance control results are obtained with full frequency-domain specifications (FFDSs). 
Considering that human movements only span the low band of the frequency domain, the frequency bandwidth requirement can be relaxed~\cite{Kong2009}.
Besides, sensor noises appear at the high-frequency band. 
Therefore, restricted frequency-domain specifications (RFDSs) can be introduced into the impedance control to further enhance the performance at the specific frequency bands. 
There are methods that introduce filters or weighting functions into the impedance control framework to indirectly improve performance in certain frequency ranges~\cite{Mehling2014,Santos2015,Yu2017AA}. 
However, there is no systematic method for design of a well-performed weighting function, and the weighting function needs to be incorporated into the augmented plant model, which increases the order of the synthesized controller. The generalized Kalman-Yakubovi\v{v}-Popov (KYP) lemma provided a possible approach to directly handle the RFDSs by converting it into equivalent linear matrix inequalities (LMIs), actually bilinear matrix inequalities (BMIs)~\cite{Iwasaki2005TAC}. But, the LMI-based approach runs into numerical difficulties due to the quadratic growth of the number of the Lyapunov variables. In~\cite{Apkarian2007,Apkarian2014}, a non-smooth optimization technique was proposed to solve the fixed-structured controller synthesis problem with multiple models, multiple objectives and multiple frequency bands. With this method, it is possible to directly synthesize a fixed-order dynamic controller to achieve multiple frequency-domain specifications for the impedance control of the SEA. 

However, stable torque control does not suffice for physical human-robot interaction, and the system has to guarantee passivity in the presence of uncertain contact dynamics~\cite{Colgate1988}. In~\cite{Vallery2008,Sergi2015,Tagliamonte2014,Calanca2017}, a symbolic and analytical method with respect to the passivity constraints has been used to derive the allowable ranges of the control parameters for the PID based impedance control structure. However, this method can not directly give the desired controller gain, and the derived symbolic inequalities with respect to all the system's parameters are very complicated. Thus, it is not suitable for structured synthesis of the controller.

In our previous work~\cite{Yu2017AA}, a mixed $H_2$/$H_\infty$ method based on a model matching framework was employed to synthesize the impedance controller with full frequency-domain specifications for a cable-driven SEA system. However, passivity could not be guaranteed.

In this paper, we address the stiffness control problem with restricted frequency-domain specifications of impedance rendering, passivity and robustness to improve the rendering performance for physical human-robot interaction with a cable-driven SEA. 
The main contributions of this paper lie in the following aspects. 
Firstly, this is the first work to formulate and realize impedance control of SEA with restricted frequency-domain specifications. Secondly, the strict passivity constraint can be guaranteed by transforming it into an equivalent norm bound over the entire frequency band. Then, a non-smooth optimization algorithm was adapted to find the solution and synthesize the controller.

The paper is organized as follows. The stiffness control problem for a cable-driven SEA and its $H_\infty$ synthesis framework are introduced in Section~\ref{section2}. Characterization of the restricted frequency-domain specifications, passivity transformation and controller synthesis are presented in Section~\ref{section3}. Extensive simulations, experiments, and results are shown in Section~\ref{section4}. Finally, Section~\ref{section5} concludes the paper.

\section{$H_{\infty}$ Formulation of the Stiffness Control Problem for A Cable-Driven SEA}
\label{section2}

\subsection{Interaction with A Cable-Driven SEA}

\begin{figure}[!htbp]
	\centering
	\includegraphics[width=0.65\columnwidth]{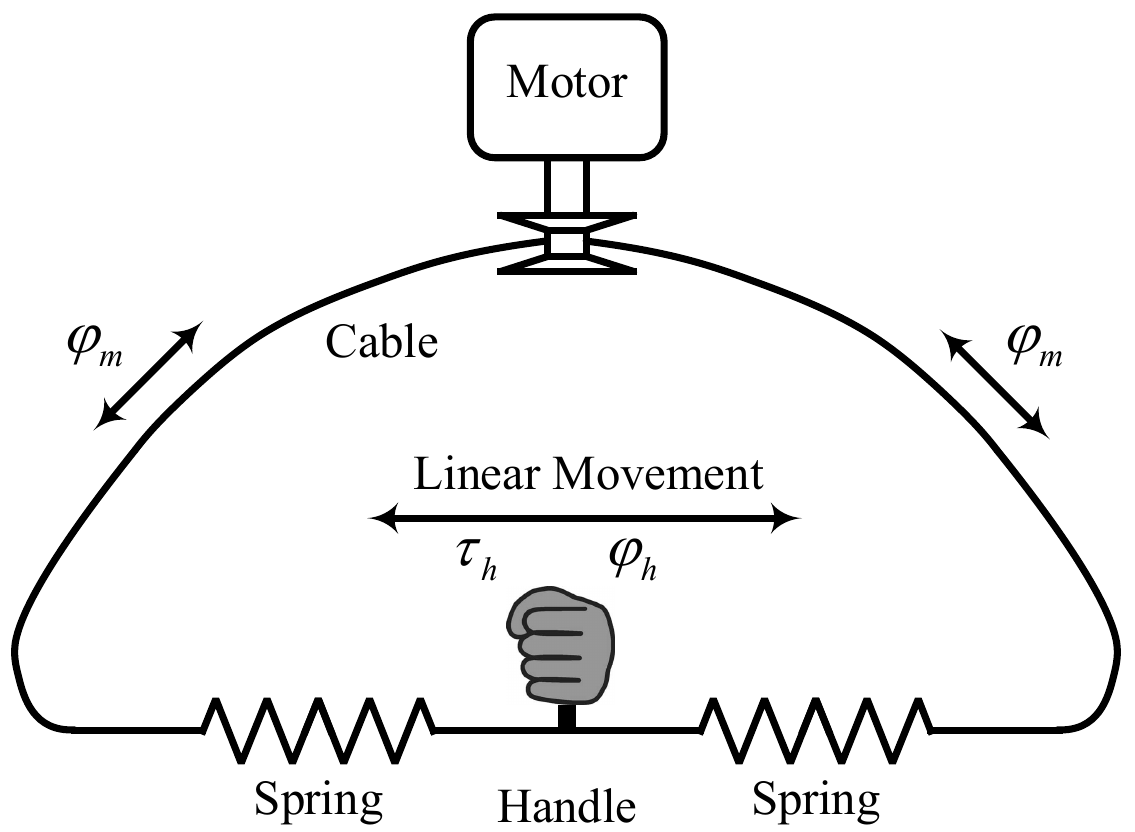}
	\caption{Human hand interacting with a cable-driven SEA.}
	\label{fig_HRI}
\end{figure}

A cable-driven series elastic actuator used for interaction with a human hand is illustrated in Fig.~\ref{fig_HRI}. The cable for force transmission and a pair of linear springs are connected in series between the driving motor and the handle. When interacting with the cable-driven SEA, the human hand drives the handle to slide along the linear guide, while the motor rotates to regulate the spring deformation that produces force. Thus, the human hand perceives the generated impedance.

We represent force and linear movements by their equivalent torque and angular variables on the motor side. The motor velocity $\omega_m$ leads to cable displacement $\varphi_m$. The two identical springs are of the spring constant $K_s$. The handle motion is represented by $\varphi_h$, and the equivalent torque acting on the human hand is $\tau_h$. Here, the deformation and inertia of the cable as well as the inertia of the springs can be neglected.

\subsection{Model Construction}

\begin{figure}[!htbp]
	\centering
	\includegraphics[width=0.9\columnwidth]{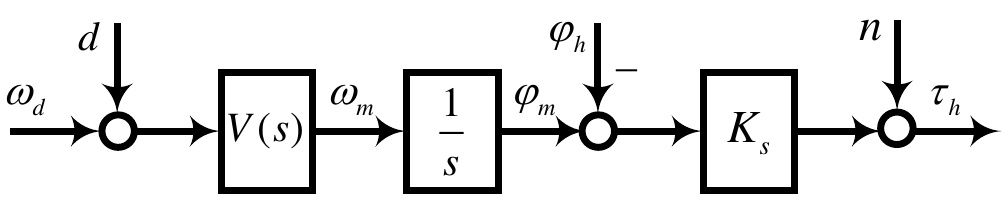}
	\caption{The linearized open-loop model of the cable-driven SEA.}
	\label{fig_velocity_sourced}
\end{figure}

The linearized model of the cable-driven SEA is depicted in Fig.~\ref{fig_velocity_sourced}. The desired motor velocity is $\omega_d$, while the motor is modeled as a velocity source with the transfer function $V(s)$ from the desired motor velocity $\omega_d$ to the actual motor velocity $\omega_m$. The disturbance $d$ and sensor noise $n$ are also considered there. A velocity controller can overcome some undesirable effects caused by motor internal disturbance.

For the above open-loop model, the transfer relation from the external inputs $\left[\varphi_h,~d,~n\right]^T$ and  control input $\omega_d$ to the output $\tau_h$ can be written as
\begin{equation}
{\tau _h} =  - {K_s}{\varphi _h} + {G_1}(s)d + n + {G_1}(s){\omega _d}.
\end{equation}
Here, we define the open-loop transfer function from $\omega_d$ to $\tau_h$ as $G_1(s)$,
\begin{equation}
G_1(s)=\frac{\tau_h(s)}{\omega_d(s)}=\frac{K_sV(s)}{s}.
\end{equation}

\subsection{$H_{\infty}$ Synthesis Formulation of Stiffness Control}

For this system, the transfer function $Z(s)$ between the torque applied to the SEA and its motion ${\varphi}_h(s)$ is defined as the impedance
\begin{equation}
Z(s)=\frac{\tau_h(s)}{-{\varphi}_h(s)}.
\end{equation}
In general cases, the impedance encompasses inertia, damping and stiffness components. 

\begin{figure*}[!ht]
	\centering
	\includegraphics[width=1.6\columnwidth]{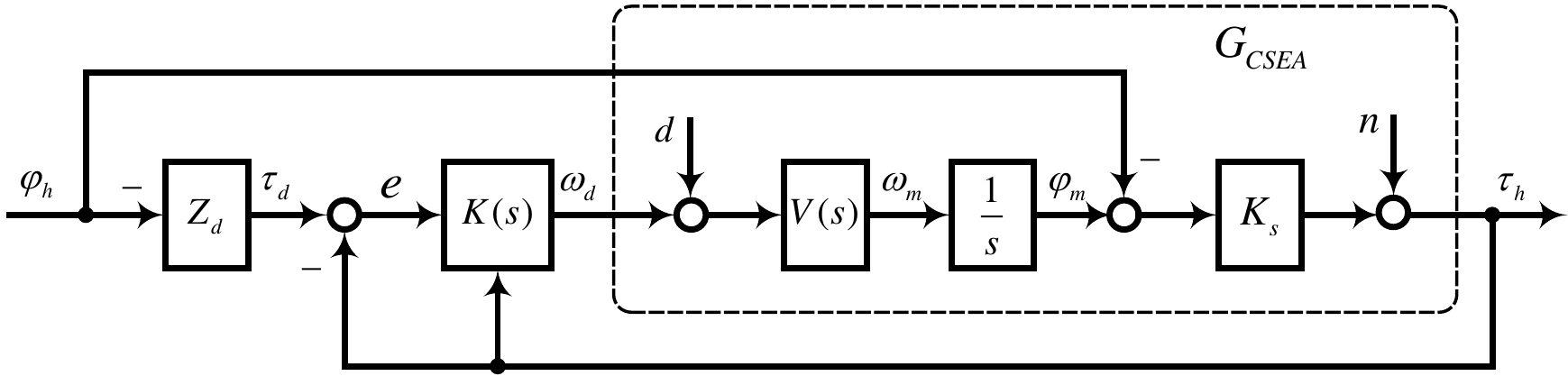}
	\caption{The general diagram of impedance control strategy for the cable-driven SEA.}
	\label{fig_impedance_control}
\end{figure*}

The goal of impedance control is to minimize the error between the actual impedance rendered by the system and the desired impedance model. 
The block diagram of the impedance control for the cable-driven SEA is depicted in Fig.~\ref{fig_impedance_control}. If the error $e$ between the desired and actual output torque is smaller, then, the actual stiffness is much closer to the desired stiffness. Minimizing the torque error is an effective means to achieve accurate impedance rendering. 

Here, $G_{CSEA}$ is the open-loop model as illustrated in Fig.~\ref{fig_velocity_sourced}. The desired impedance $Z_d$ takes pure stiffness proportional to the physical stiffness $K_s$ in this work. The desired torque is denoted by $\tau_d$. The exogenous input vector is $w=\left[\varphi_h,~d,~n\right]^T$, and the controller output is $u=\omega_d$. A structured dynamic output-feedback controller $K(s)=\left[K_1(s),~K_2(s)\right]$ will be designed here, with a given order $n_k$. The feedback signal is the vector of measured outputs $y=\left[\tau_h,~e\right]^T$. Then, $z=\left[\tilde{e},~u\right]^T$ is the output vector of interest to be optimized, representing the weighted error and control effort. 

Thus, the impedance control problem has been formulated into the $H_{\infty}$ synthesis framework: to find a stabilizing dynamic output-feedback controller $K(s)$ such that the magnitude of the transfer function $T_{w_i{z_i}}(s)$ from exogenous input $w_i$ to target output $z_i$ is bounded by a small scalar $\gamma_i>0$ within a certain frequency range $\Omega_i$,
\begin{equation} \label{eq_H_infty_formulation}
\left| {{T_{w_i{z_i}}}(j\omega )} \right| \le {\gamma _i},~\omega  \in {\Omega _i}.
\end{equation}

This $H_{\infty}$ synthesis formulation is a great advance from the conventional impedance control approaches. System properties and control limitations can be concretely analyzed. Constraints on impedance rendering, passivity, control effort, noise rejection and disturbance attenuation, etc, can be incorporated into this framework with their relative frequency bands to conduct multi-objective, multi- frequency-domain optimization.

\textbf{Remark 1.} For the structured controller design, the designer can pre-define the controller structure, such as the order $n_k$, degrees of freedom and proper measurements as the controller inputs.

\textbf{Remark 2.} To eliminate system singularity and guarantee that the formulated $H_\infty$ synthesis problem is well-posed, a frequency-domain weighting function $W_e(s)$ is designed for the error $e$, and the weighted error is $\tilde{e}=W_ee$. 
It is noteworthy that, unlike the model matching framework in our previous work~\cite{Yu2017AA} in which each signal should be appended with a weighting function to adjust the performance, only one weighting function $W_e(s)$ is needed in this work to eliminate system singularity.

\textbf{Remark 3.} The non-smooth optimization method used in this work (to be presented in Section \ref{sec_synthesis}) is able to synthesize a desired controller even when the weighting function $W_e(s)$ is not used. Nevertheless, the optimization process is much faster with the weighting function $W_e(s)$ to ensure the synthesis problem is well-posed. 

\textbf{Remark 4.} To our knowledge, there is no existing work introduced restricted frequency-domain constraints for the SEA impedance control. This is the first work to formulate and realize impedance control with restricted frequency-domain specifications.

\section{Performance Characterization and Controller Synthesis}
\label{section3}

\subsection{Characterization of Passivity by Equivalent Transformation}
\label{Section_Passivity}

For a system interacting with human, passivity should be ensured. We define $\bar{Z}(s)$ as the transfer function from the interaction velocity $\dot{\varphi}_h$ to the interaction torque applied to the mechanical system, that is
\begin{equation}
\label{eq_passivity}
\bar Z(s) = \frac{{{\tau _h}(s)}}{{ - {{\dot \varphi }_h}(s)}} = \frac{Z(s)}{s}.
\end{equation}
Then, there are two necessary and sufficient conditions for $\bar{Z}(s)$, which should be satisfied to guarantee system passivity~\cite{Colgate1988,Vallery2008}.
\begin{itemize}
	\item $\bar Z(s)$ must be stable such that $\bar Z(s)$ has no poles in the right-half of the complex $s$-plane.
	\item The real part of $\bar Z(j\omega)$ must be nonnegative for all $\omega$ for which $j\omega$ is not the pole of $\bar Z(s)$.
\end{itemize}

The two conditions can provide us a symbolic and analytical approach to derive the allowable ranges of the control parameters~\cite{Vallery2008,Sergi2015,Tagliamonte2014,Calanca2017}. However, this method can not directly give the controller gain for the desired performance, and the symbolic expressions with respect to all the system's parameters are much complex. Thus, this approach is not suitable for the stiffness controller synthesis framework. A more straightforward and efficient way is demanded here to meet the passivity constraint.

\begin{figure}[!htbp]
	\centering
	\includegraphics[width=0.5\columnwidth]{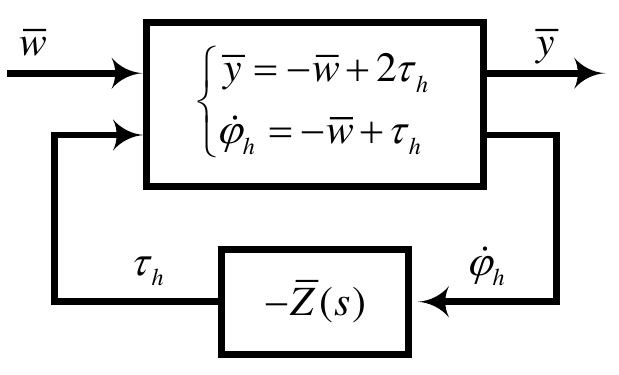}
	\caption{The equivalent transformation for the passivity constraint.}
	\label{fig_passivity_LFT}
\end{figure}

In~\cite{Anderson1972}, Anderson pointed out that the passivity theorem is equivalent to the small gain theorem by the transformation shown in Fig.~\ref{fig_passivity_LFT}. 

\begin{equation}
\label{eq_req_passivity}
\left| {\frac{{\bar Z(j\omega)}-1}{{\bar Z(j\omega)}+1}} \right|=\left| T_{\bar{w}\bar{y}}(j\omega) \right| \le 1,~\omega  \in R.
\end{equation}

If (\ref{eq_req_passivity}) is satisfied, then, the impedance $\bar{Z}(s)$ is passive. To achieve a global passivity, the $H_\infty$ norm of the transfer function $T_{\bar{w}\bar{y}}(s)$ from the imaginary input $\bar w$ to the imaginary output $\bar y$ should be bounded by 1.

This transformation enables us to incorporate the passivity constraint into the $H_\infty$ synthesis framework and makes it possible to design a controller that directly guarantees passivity of the closed loop system.

\subsection{Performance Characterization with Restricted Frequency-Domain Specifications}

\subsubsection{Characterization of Impedance Rendering Requirements}
Physical human-robot interaction mostly happens in a certain low frequency band, because both human limbs and robotic actuators can only produce force/torque or movement with limited bandwidth~\cite{Yu2017AA,Kong2009}. Optimizing the performance across the full frequency domain may produce conservative controllers. In this work, we impose frequency-domain specifications to enhance the impedance control performance in the relative desired frequency bands. The impedance rendering error just needs to be minimized or bounded over a specified low frequency band. For the transfer function from the motion $\varphi_h$ to the weighted torque error $\tilde{e}$, the restricted frequency-domain specification is
\begin{equation}
\label{eq_req_e}
\left| {{T_{{\varphi _h}\tilde e}}(j\omega )} \right| \le {\gamma _1},~\left|\omega\right|  \le {\omega _e}.
\end{equation}

Here, $\left[0,~\omega_e\right]$ characterizes the low frequency range for physical human-robot interaction. This requirement specifies the maximal gain of the response $T_{\varphi_h{\tilde{e}}}(j\omega)$ should not exceed $\gamma_1$ at the specified low frequency range.

\subsubsection{Characterization of Actuator Limitations}

We further consider the control effort and motor saturation. To generate a reasonable controller output $u$ under the actuator saturation limit, we impose the following restricted frequency-domain specification to constrain the transfer function from the motion $\varphi_h$ to the controller output $u$.
\begin{equation}
\label{eq_req_u}
\left| {{T_{{\varphi _h}u}}(j\omega )} \right| \le {\gamma _2},~\left|\omega\right| \le {\omega _u}.
\end{equation}

Here, $\left[0,~\omega_u\right]$ characterizes the low frequency range for the controller output.

\subsubsection{Characterization of Disturbance Attenuation}

The external disturbances and parameter uncertainties in this system are all included in the combined disturbance $d$ acting on the plant input. Since it may span all the frequency range, the following constraint can be imposed to the transfer function from the disturbance $d$ to the interaction torque $\tau_h$, such that
\begin{equation}
\label{eq_req_d}
\left| {{T_{d{\tau _h}}}(j\omega )} \right| \le {\gamma _3},~\omega  \in R.
\end{equation}

Of course, to limit the response from the disturbance $d$ to the controller output $u$, one can take 
\begin{equation}
\label{eq_req_du}
\left| {{T_{du}}(j\omega )} \right| \le {\gamma _4},~\omega  \in R.
\end{equation}

\subsubsection{Characterization of Noise Rejection}

The noise in the cable-driven SEA system results from the displacement sensors, whose frequency range is largely dominated by system's sampling frequency. Thus, a restricted frequency-domain specification for noise rejection in the high frequency domain is imposed to the transfer function from the noise $n$ to the interaction torque $\tau_h$:
\begin{equation}
\label{eq_req_n}
\left| {{T_{n{\tau _h}}}(j\omega )} \right| \le {\gamma _5},~\left|\omega\right| \ge {\omega _n}.
\end{equation}

Here, $\left[\omega_n,~\infty\right)$ characterizes the high frequency range of the noise. 

Also, to limit the response from the noise $n$ to the controller output $u$, one can take 
\begin{equation}
\label{eq_req_nu}
\left| {{T_{nu}}(j\omega )} \right| \le {\gamma _6},~\left|\omega\right| \ge {\omega _n}.
\end{equation}

\subsection{Multi- Frequency-Band Stiffness Controller Synthesis}
\label{sec_synthesis}

\begin{figure}[!htbp]
	\centering
	\includegraphics[width=0.85\columnwidth]{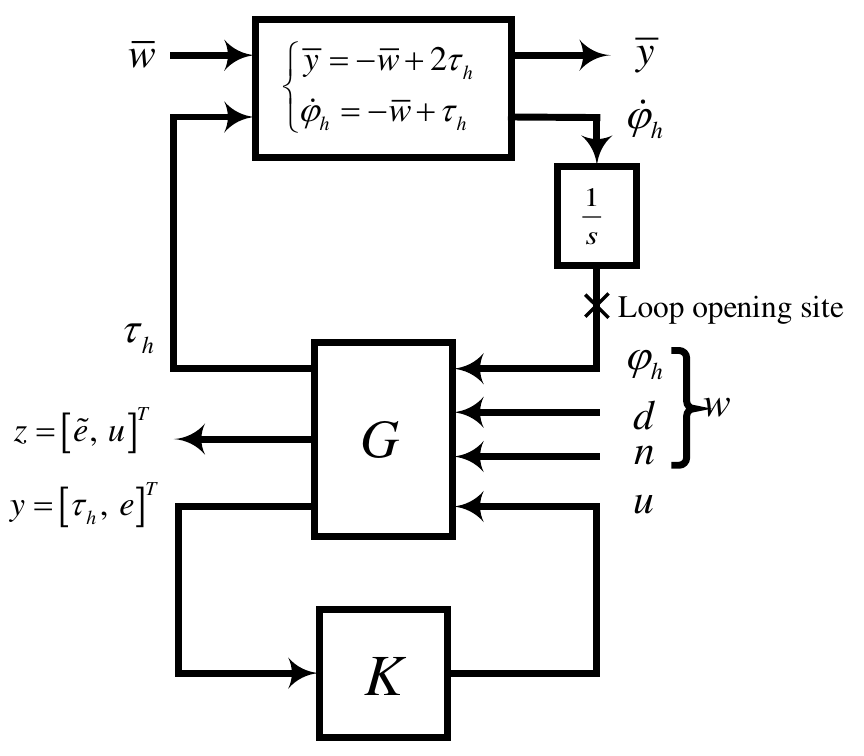}
	\caption{Controller synthesis with multiple performance requirements.}
	\label{fig_impedance_control_multiple}
\end{figure}

The stiffness controller synthesis with multiple performance requirements is illustrated in Fig.~\ref{fig_impedance_control_multiple}. $G$ represents the open-loop transfer relation of the impedance control framework in Fig.~\ref{fig_impedance_control}. The goal is to synthesize a stabilizing dynamic controller $K(s)$ to meet the restricted frequency-domain requirements in (\ref{eq_req_passivity}), (\ref{eq_req_e}), (\ref{eq_req_u}), (\ref{eq_req_d}), (\ref{eq_req_du}), (\ref{eq_req_n}), (\ref{eq_req_nu}) simultaneously as much as possible.

However, as shown in Fig.~\ref{fig_impedance_control_multiple}, the model of the passivity constraint is different from the others. The loop opening site should be unconnected when coping with the requirements (\ref{eq_req_e})-(\ref{eq_req_nu}), while it needs to be closed when considering passivity. This configuration (some loops are open while others are closed) makes the synthesis problem extremely tough for the conventional LMI-based $H_\infty$ synthesis methods.

In this work, we adapt a non-smooth optimization algorithm~\cite{Apkarian2007,Apkarian2014} to cope with such a multi-model, multi-objective, multi- frequency-band controller synthesis problem. This method can avoid the difficulties introduced by the LMIs and specific configuration with some loops open while others closed.

\section{Simulation and Experiment Results}
\label{section4}

\subsection{Experimental Setup}

\begin{figure}[!ht]
	\centering
	\includegraphics[width=0.85\columnwidth]{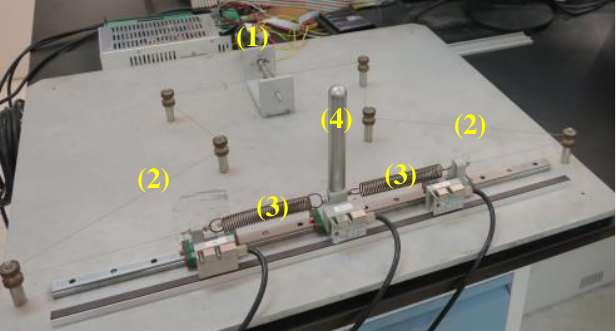}
	\caption{The cable-driven SEA platform. (1) the DC motor; (2) cable; (3) the linear tension spring; (4) interactive handle.}
	\label{fig_prototype}
\end{figure}

The prototype of the cable-driven SEA for physical human-robot interaction is shown in Fig.~\ref{fig_prototype}. Its main parameters are shown in Table~\ref{table1}.

\begin{table}[!htbp]
	\caption{Main parameters of the cable-driven SEA platform.}
	\label{table1}
	\begin{center}
		\renewcommand{\arraystretch}{1.6}
		\begin{tabular}{ll}
			\hline \hline		
			Stiffness of double springs  & 920 N/m\\
			\hline
			Equivalent rotational stiffness $K_s$ & 0.0484 Nm/rad\\
			\hline
			Radius of cable winch & 7.25 mm\\
			\hline
			Ratio of gear head & 14:1\\
			\hline
			Motor saturation & 44 rad/s\\
			\hline \hline
		\end{tabular}
	\end{center}
\end{table}

The transfer function $G_1(s)$ from the desired velocity $\omega_d$ to the interaction torque $\tau_h$ was identified as
\begin{equation}
\begin{aligned}
G_1(s)&=\displaystyle \frac{\tau_h(s)}{\omega_d(s)}=\displaystyle -\frac{0.1064 s + 279.4}{s^3 + 81.64 s^2 + 5821 s + 1389}\\
&= - \frac{{2.200s + 5778}}{{{s^2} + 81.44s + 5802}} \times \frac{1}{{s + 0.2394}} \times 0.04840
\end{aligned}.
\end{equation}
The velocity-controlled motor was identified as a  second-order transfer function $V(s)$. Because of the motor saturation and other physical limits, the pure integration in $G_1(s)$ was replaced by a first-order filter with a small pole $-0.2394$.

The weighting function $W_e(s)$ can guarantee the synthesis problem being well-posed, and largely accelerate the optimization process. It was designed with respect to the sensitivity requirement~\cite{ZhouKemin1998}, and the detailed design procedure for the cable-driven SEA system can be found in our previous work~\cite{Yu2017AA}.

We mainly considered the frequency-domain requirements in equations (\ref{eq_req_passivity}), (\ref{eq_req_e}), (\ref{eq_req_u}), (\ref{eq_req_d}) and (\ref{eq_req_n}) in this work.
The velocity saturation of the DC motor was 44 rad/s, thus, the bound to the controller output $u$ was set as $\gamma_2=44$.
The static gain of the open-loop transfer function from disturbance $d$ to torque $\tau_h$ was $\left| {{G_1}(0)} \right| \approx 0.20$. The parameter $\gamma_3$ was set as $\gamma_3=0.03$ so that the maximal gain of the closed-loop response $T_{d\tau_h}(s)$ did not exceed $0.03$ (about $-30$ dB). 
The static gain of the open-loop transfer function from noise $n$ to torque $\tau_h$ was 1. The parameter $\gamma_5$ was set as $\gamma_5=0.3$ so that the maximal gain of the closed-loop response $T_{n\tau_h}(s)$ did not exceed $0.3$ (about $-10$ dB) at the specified high frequency range.
Since human hand motion commonly falls in the low frequency range, i.e., from 0 to 6 Hz, the parameters $\omega_e$ and $\omega_u$ were set as $12\pi$ rad/s (6 Hz). The parameter $\omega_n$ was set as $40\pi$ rad/s (20 Hz) to reject high-frequency noise. The controller order $n_k$ was set as 3, equal to the order of $G_1(s)$.

The error bound for impedance rendering, $\gamma_1$, was adjusted manually, and it depends on the desired impedance. For a specified $Z_d$, a smaller bound $\gamma_1$ guarantees more accurate impedance rendering. However, if $\gamma_1$ is too small, the optimization method may fail to synthesize a feasible controller. When the desired impedance $Z_d$ is close to the physical stiffness $K_s$,  $\gamma_1$ can be set close to zero. While $Z_d$ is close to 0 (zero impedance control), which is challenging to render, $\gamma_1$ should be enlarged to relax the error bound for impedance rendering.

\subsection{Simulations and Results}

For $Z_d=0.6K_s$, the bound $\gamma_1$ is set as $\gamma_1=0.016$, and the synthesized controller $K(s)=\left[K_1(s),\,K_2(s)\right]$ satisfying all the requirements  (\ref{eq_req_passivity}), (\ref{eq_req_e}), (\ref{eq_req_u}), (\ref{eq_req_d}) and (\ref{eq_req_n}) is
\begin{equation}
\begin{aligned}
&{K_1}(s) =\displaystyle \frac{{4953 s^3 + 2.925e07 s^2 + 4.346e10 s - 8.294e09}}{{s^3 + 3208 s^2 + 8.408e06 s + 8.026e09}},\\
&{K_2}(s) =\displaystyle \frac{{-2180 s^3 - 7.023e06 s^2 - 1.984e09 s - 7.179e11}}{{s^3 + 1444 s^2 + 2.133e06 s + 4.533e08}}
\end{aligned}
\end{equation}

\begin{figure*}[!ht]
	\centering
	\subfigure[]{\includegraphics[width=0.83\columnwidth]{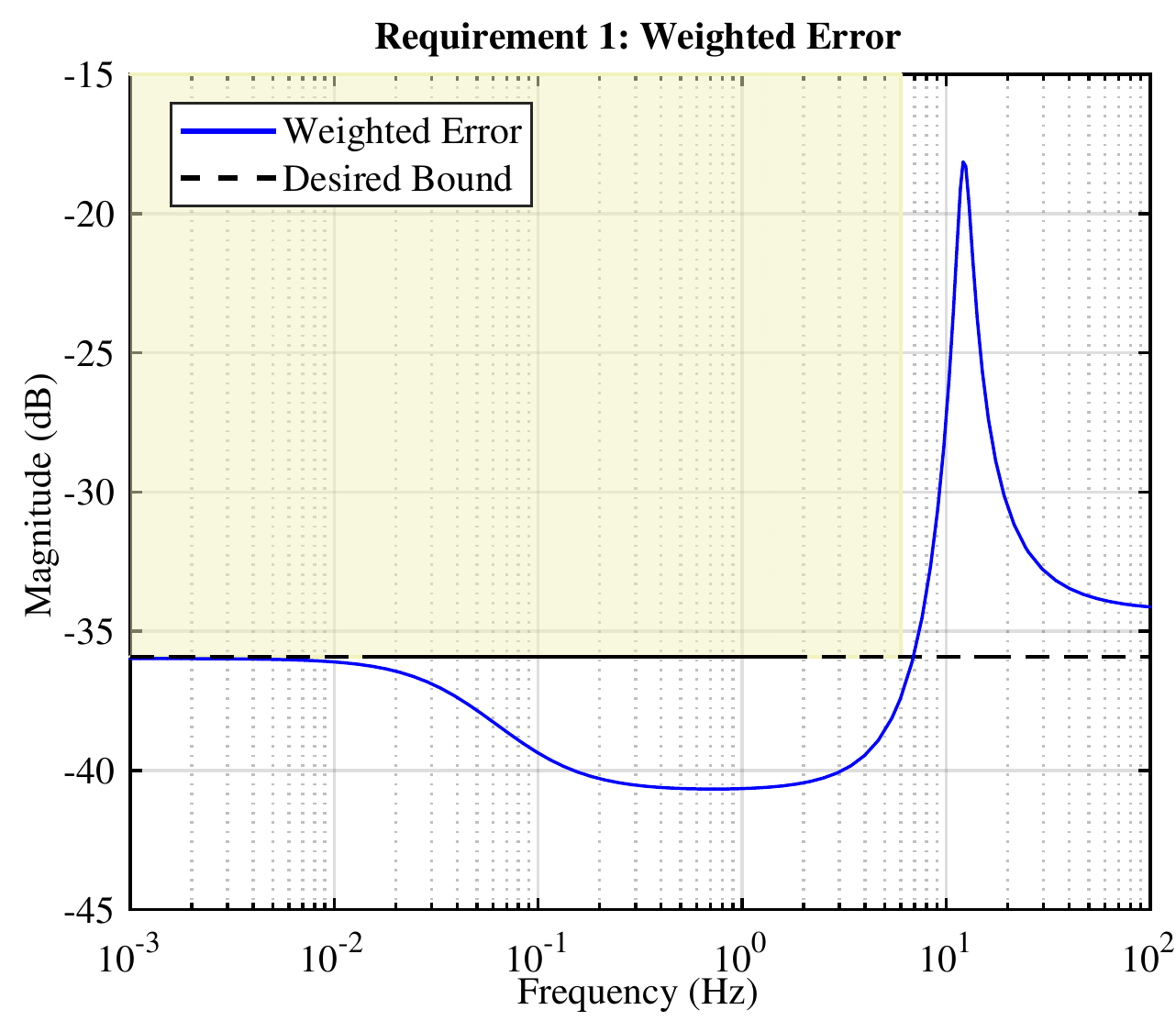}}\quad\quad\quad
	\subfigure[]{\includegraphics[width=0.83\columnwidth]{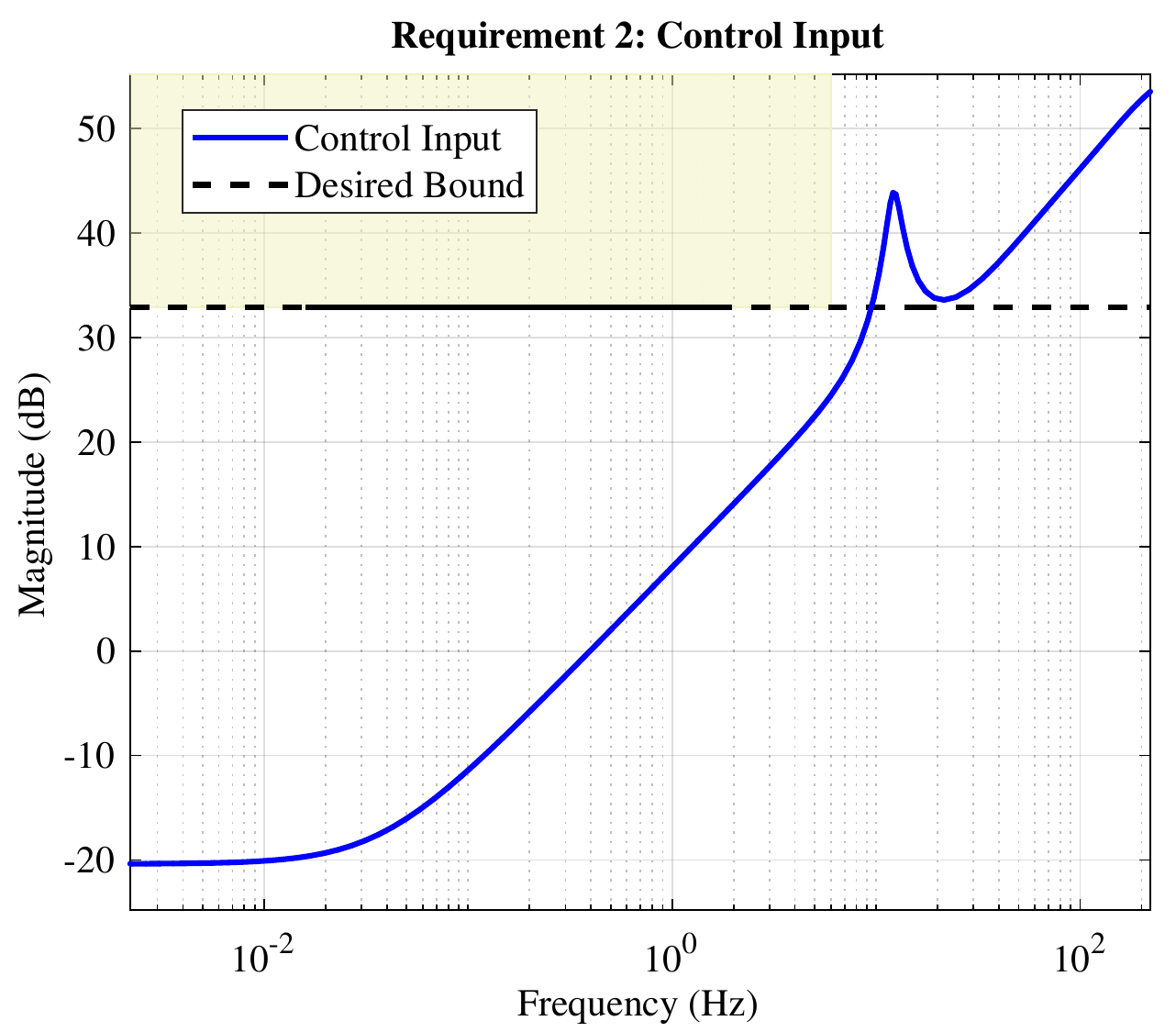}}\\
	\subfigure[]{\includegraphics[width=0.83\columnwidth]{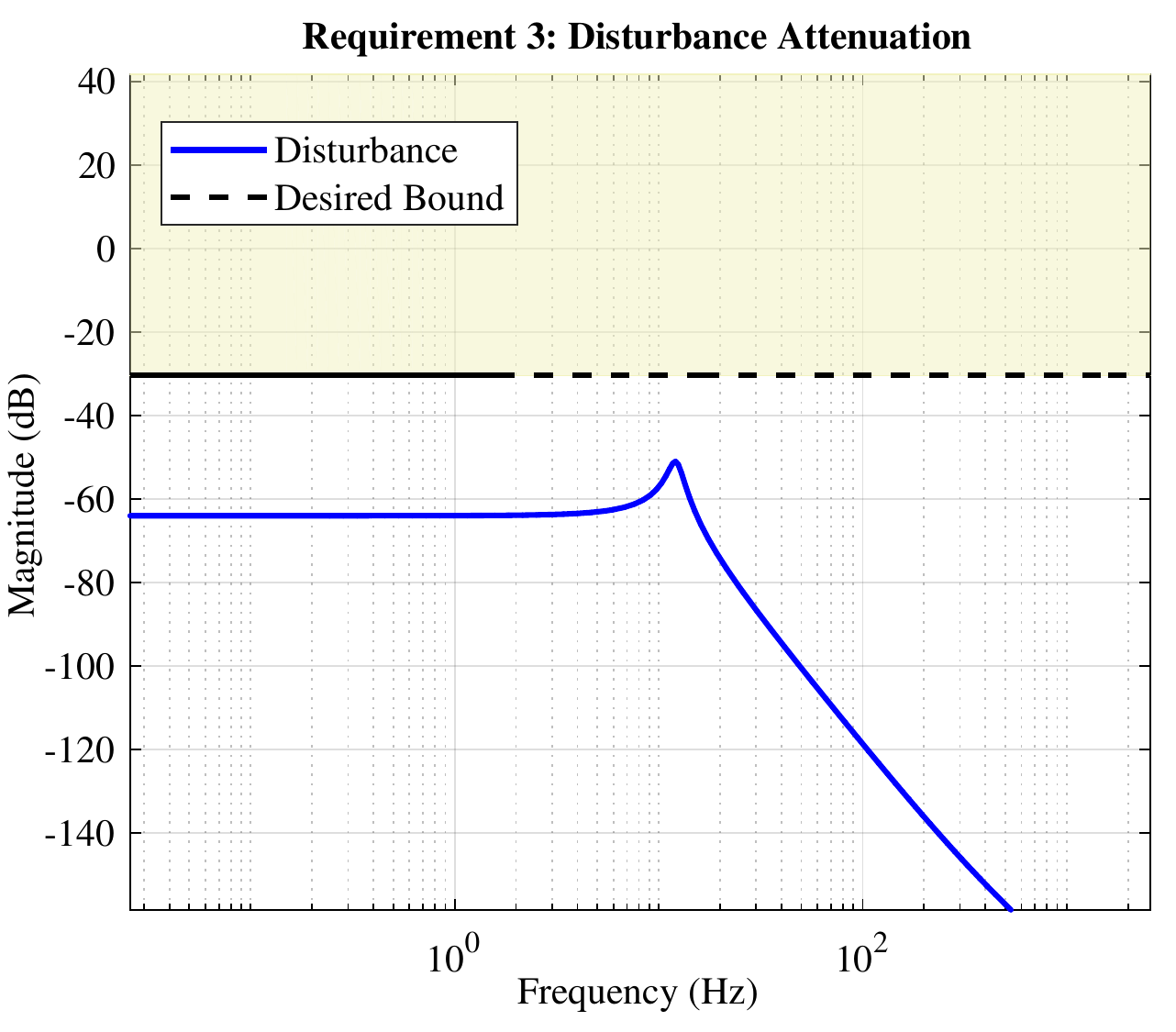}}\quad\quad\quad
	\subfigure[]{\includegraphics[width=0.83\columnwidth]{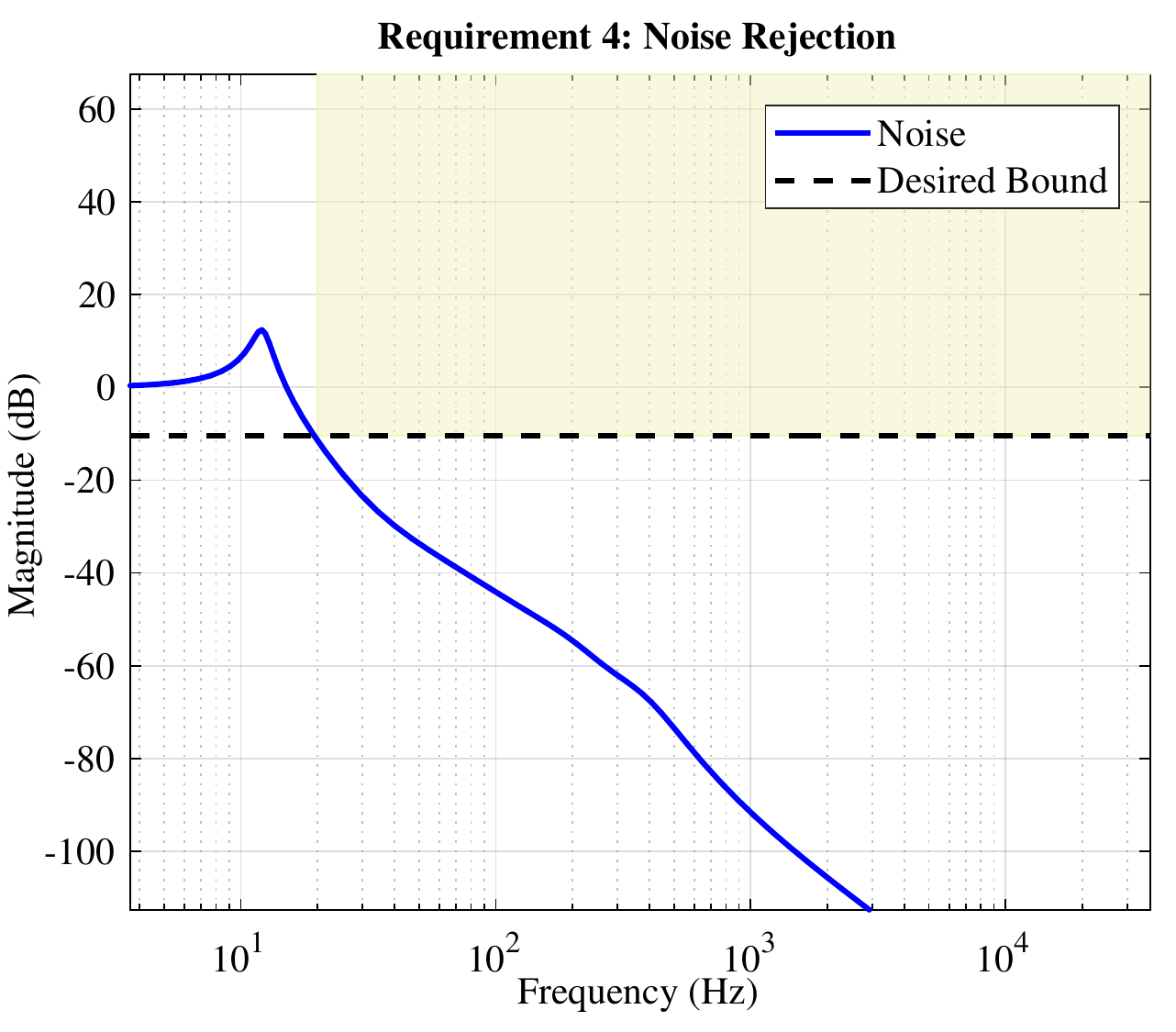}}\quad\quad\quad
	\subfigure[]{\includegraphics[width=0.83\columnwidth]{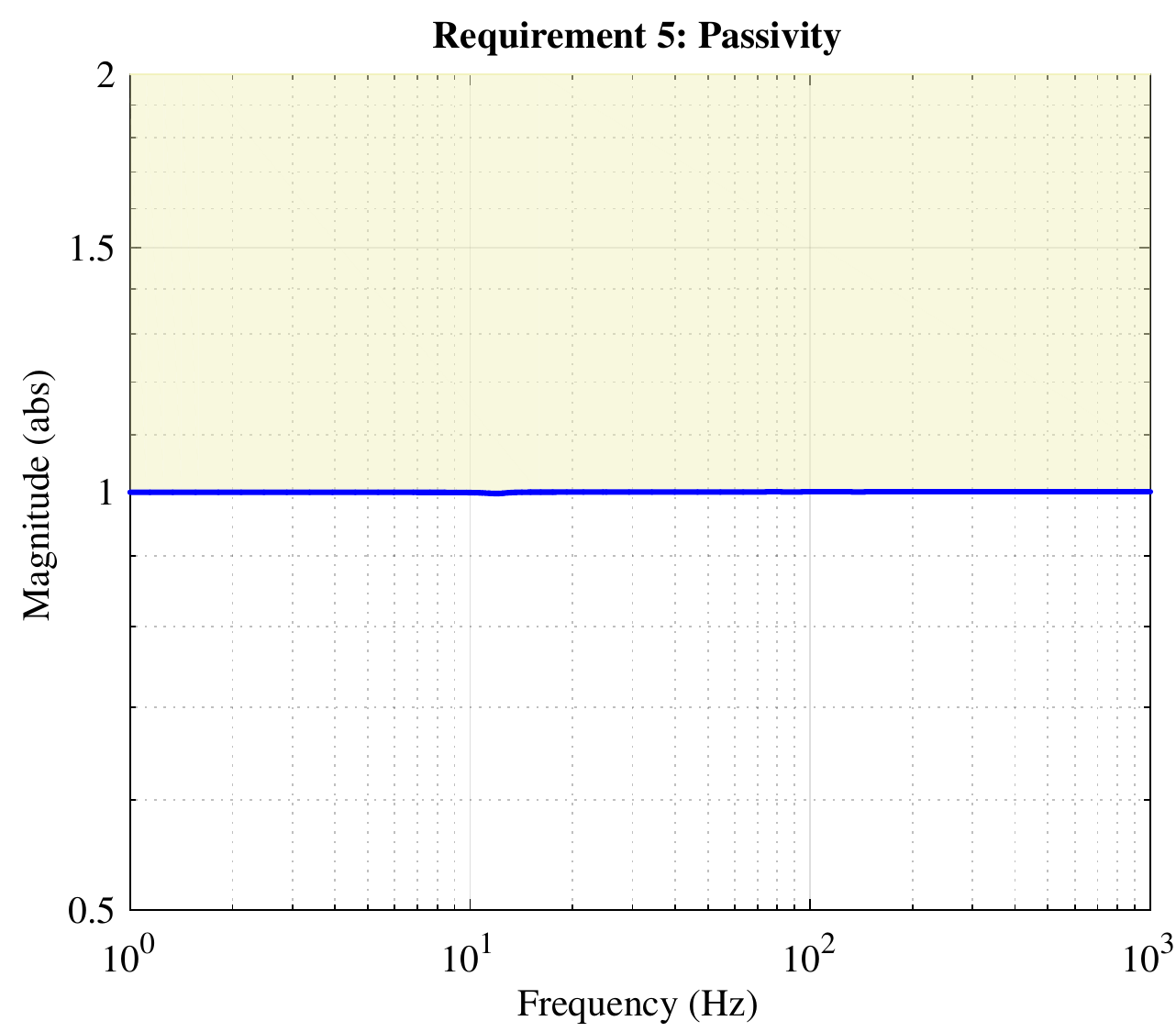}}\quad\quad\quad
	\subfigure[]{\includegraphics[width=0.83\columnwidth]{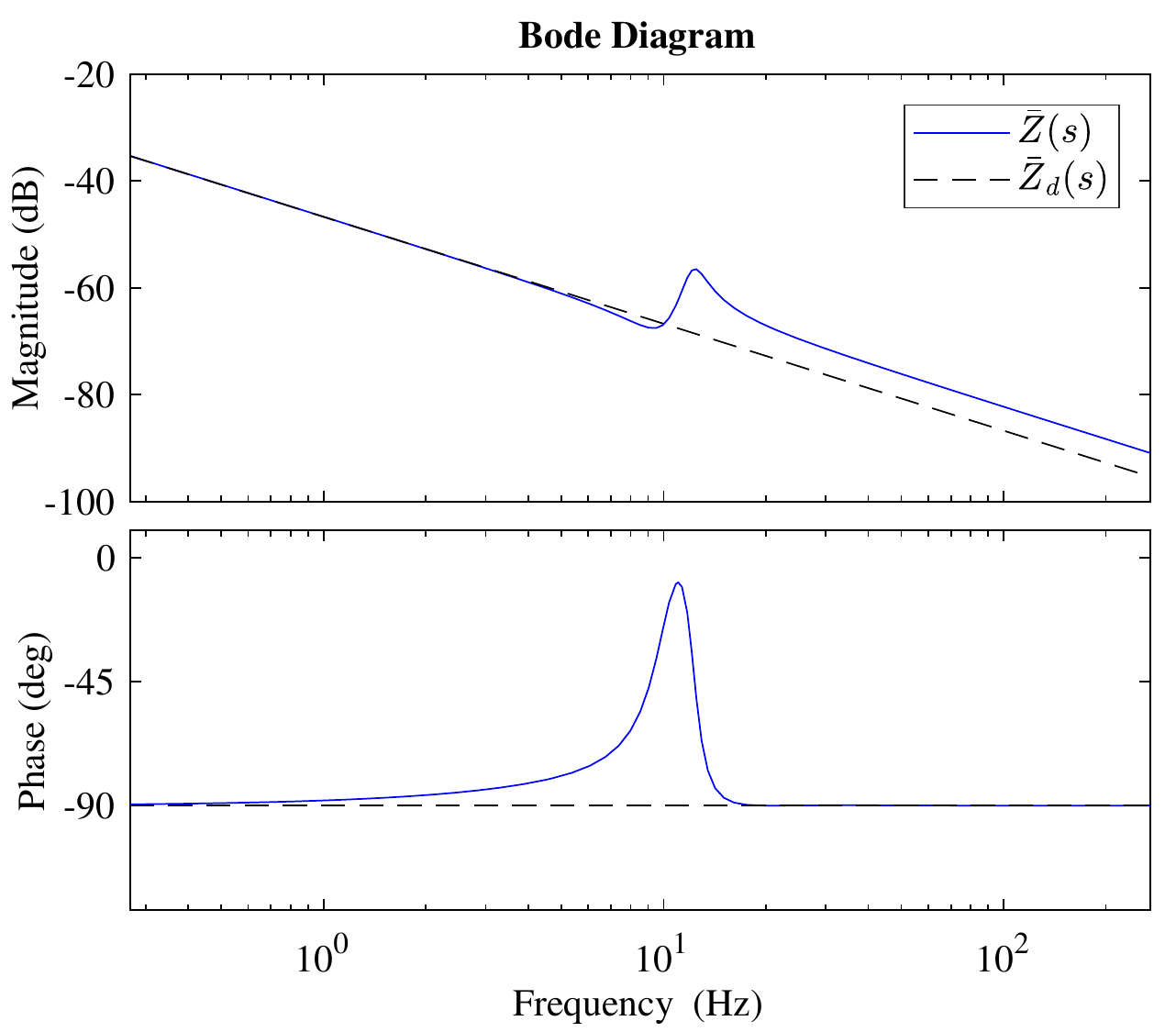}}
	\caption{Simulated frequency responses of requirements (\ref{eq_req_passivity}), (\ref{eq_req_e}), (\ref{eq_req_u}), (\ref{eq_req_d}) and (\ref{eq_req_n}) for $Z_d=0.6K_s$ for $H_\infty$ synthesis-based method (3rd-order controller). (a) Response of $\left| {{T_{{\varphi _h}\tilde e}}(j\omega )} \right|$ with $\gamma_1=0.016$, $\omega_e=6$ Hz. (b) Response of $\left| {{T_{{\varphi _h}u}}(j\omega )} \right|$ with $\gamma _2=44$ and $\omega _u=6$ Hz. (c) Response of $\left| {{T_{d{\tau _h}}}(j\omega )} \right|$ with $\gamma _3=0.03$. (d) Response of $\left| {{T_{n{\tau _h}}}(j\omega )} \right|$ with $\gamma _5=0.3$ and $\omega _n=20$ Hz. (e) Response of $\left| {\frac{{\bar Z(j\omega)}-1}{{\bar Z(j\omega)}+1}} \right|$. (f) Bode plots of the actual impedance $\bar{Z}(s)$ and the desired impedance $\bar{Z}_d(s)=\frac{0.6K_s}{s}$.}
	\label{fig_Simulation_Responses}
\end{figure*}

The frequency response of $T_{\varphi_h\tilde{e}}(s)$ is depicted in Fig.~\ref{fig_Simulation_Responses}(a). Obviously, the weighted impedance rendering error was bounded by $\gamma_1=0.016$ at the specified low frequency range. The system would achieve good stiffness rendering effect with low control error when interacting with the human hand.

During the interaction, the generated controller output $u$ should not become very extreme. The frequency response of $T_{\varphi_hu}(s)$ illustrated in Fig.~\ref{fig_Simulation_Responses}(b) showed that the controller output $u$ was also bounded by the saturation limit $\gamma_2=44$ at the specified low frequency range.

The frequency responses of $T_{d\tau_h}(s)$ in Fig.~\ref{fig_Simulation_Responses}(c) and $T_{n\tau_h}(s)$ in Fig.~\ref{fig_Simulation_Responses}(d) were all bounded within the specified frequency ranges respectively, which implied that the system will have good disturbance attenuation and noise rejection effect.

The response of $T_{\bar{w}\bar{y}}(s)$ was drawn in Fig.~\ref{fig_Simulation_Responses}(e) to show the passivity property of the interaction system. The $H_\infty$ norm of $T_{\bar{w}\bar{y}}(s)$ was bounded by 1 (0 dB), which implied the system should be passive. The passivity constraint was also equivalent to bounding the phase of $\bar{Z}(s)$ to the range of $\left[-90^\circ,\,90^\circ\right]$ across the entire frequency band. The bode plots of the closed-loop system's impedance $\bar{Z}(s)$ and the desired impedance $\bar{Z}_d(s)$ were depicted in Fig.~\ref{fig_Simulation_Responses}(f). The phase of the actual impedance $\bar{Z}(s)$ was within the range of $\left[-90^\circ,\,90^\circ\right]$. Thus, the system can maintain stable interaction under the external dynamical motion input. Besides, less discrepancy between the magnitude plots of the desired and the actual impedance at low frequency range indicated good rendering accuracy.

\begin{figure}[!htbp]
	\centering
	\includegraphics[width=0.9\columnwidth]{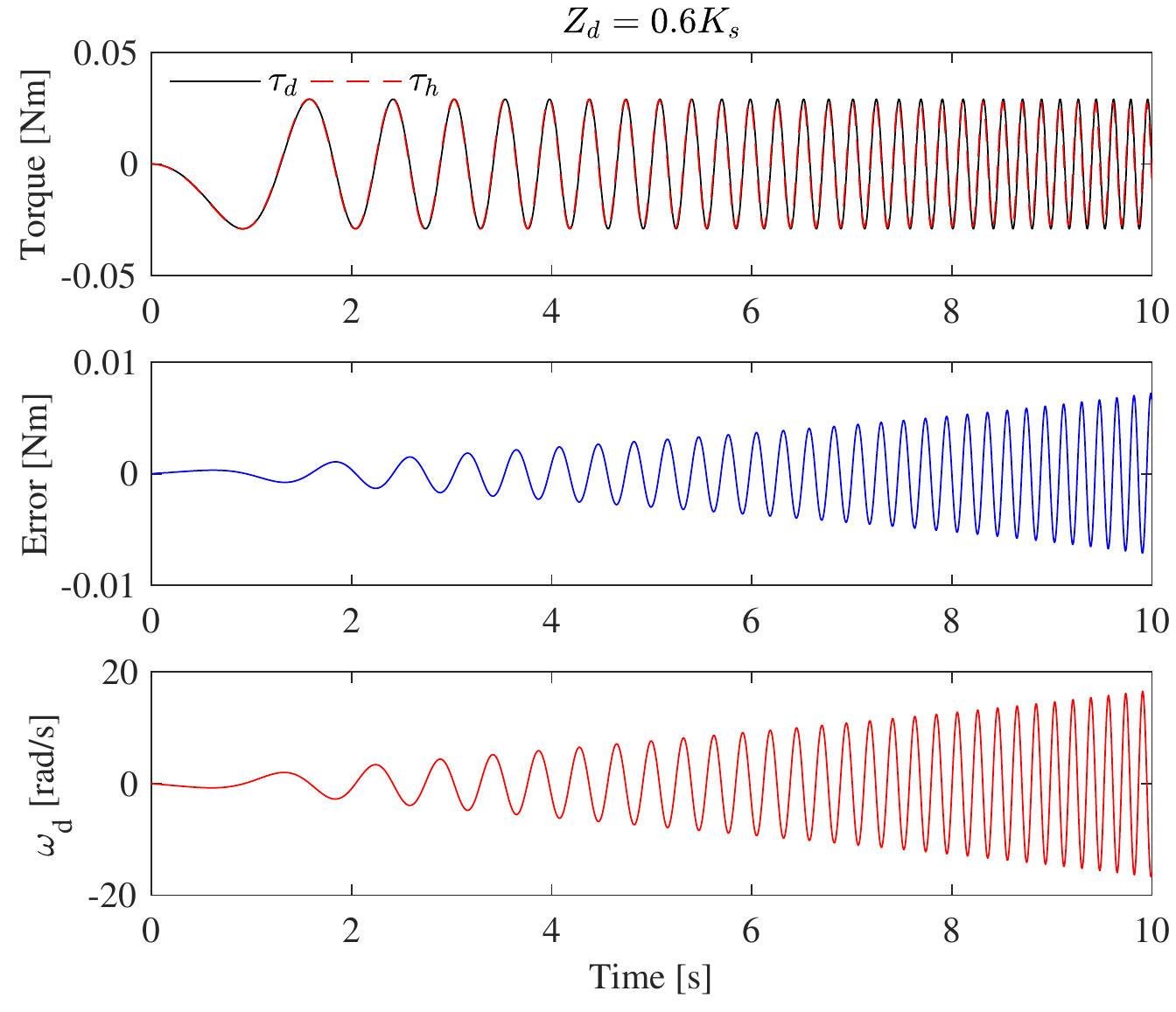}
	\caption{Simulation results of $H_\infty$ synthesis-based stiffness control (3rd-order controller) for $Z_d=0.6K_s$ when the motion $\varphi_h$ varying from 0 to 6 Hz.}\label{fig_Simulation}
\end{figure}

To illustrate the rendering performance at different frequencies, simulation results of stiffness control for $Z_d=0.6K_s$ were presented in Fig.~\ref{fig_Simulation}. The human hand motion $\varphi_h$ was set as a chirp signal varying from 0 to 6 Hz. The actual interaction torque $\tau_h$ tracked well the desired torque $\tau_d$ with a maximal error of 0.0072 Nm, and the maximal controller output $\omega_d$ was 16.7 rad/s.

\subsection{Experiments and Results}

Several experiments with different desired stiffness were conducted with the cable-driven SEA platform. In each stiffness rendering case, the human hand drove the handle to move along the linear guide with different frequencies and magnitudes.

\begin{figure*}[!ht]
	\centering
	\subfigure[]{\includegraphics[width=0.82\columnwidth]{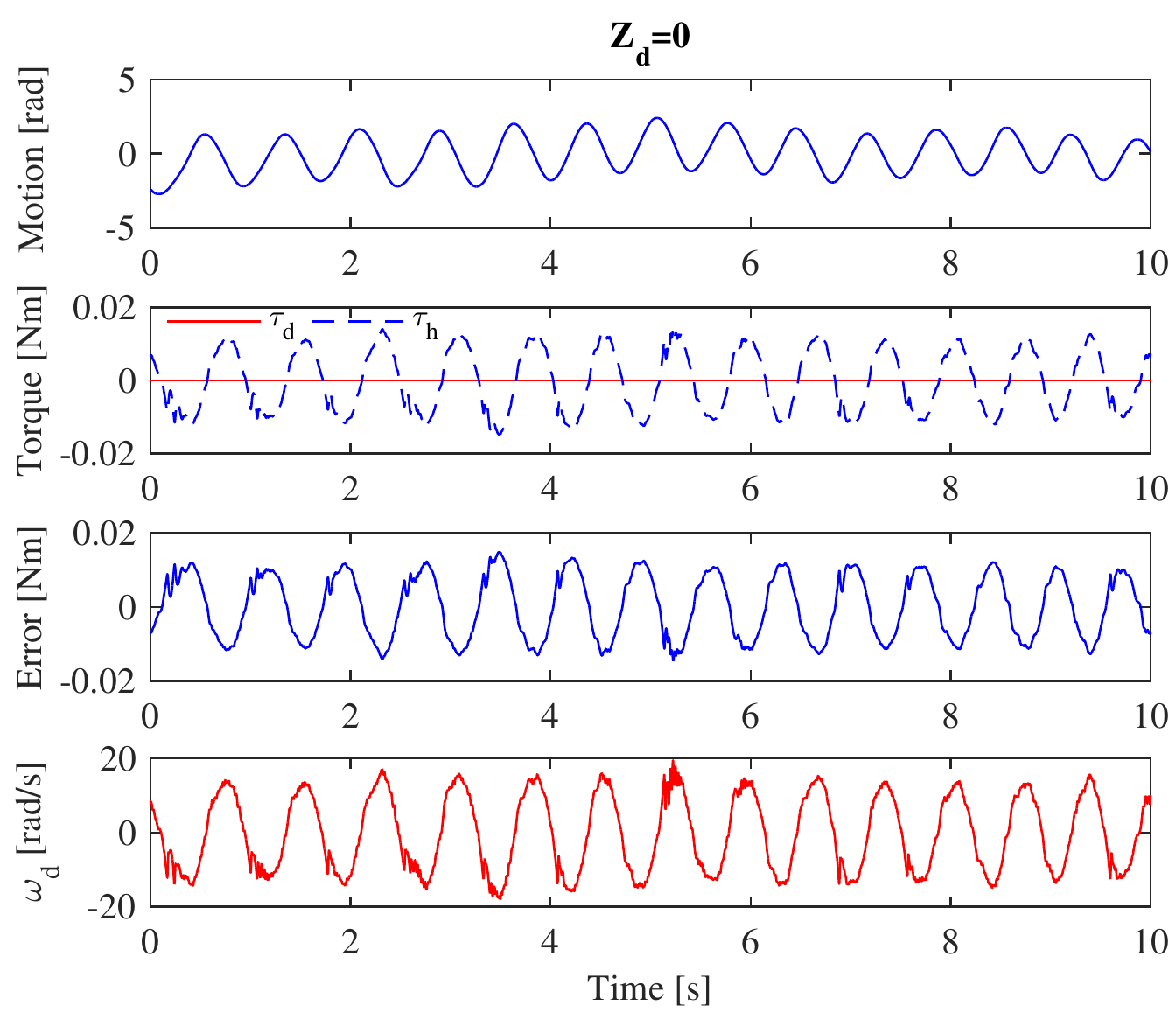}}\quad\quad\quad
	\subfigure[]{\includegraphics[width=0.82\columnwidth]{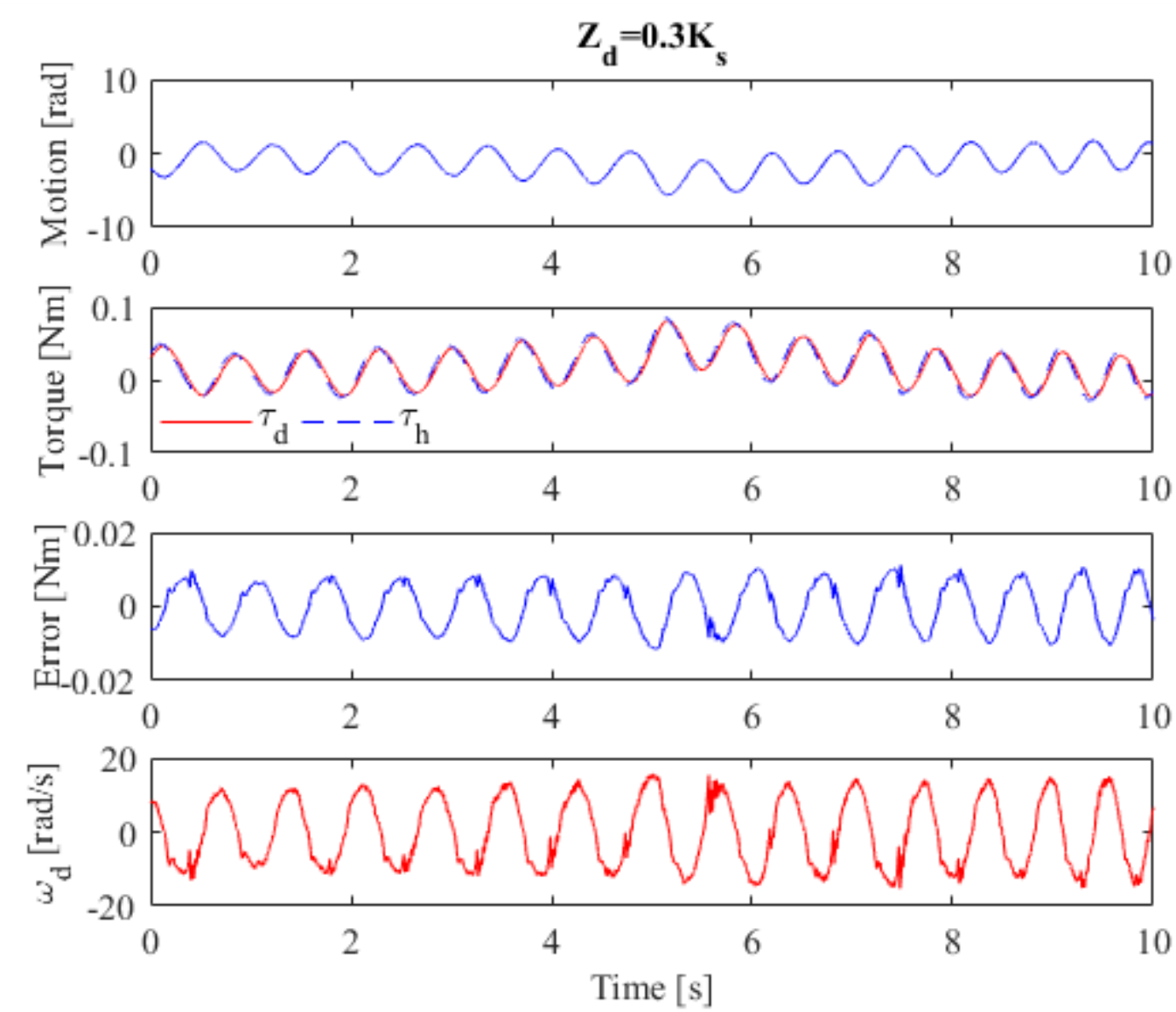}}\\
	\subfigure[]{\includegraphics[width=0.82\columnwidth]{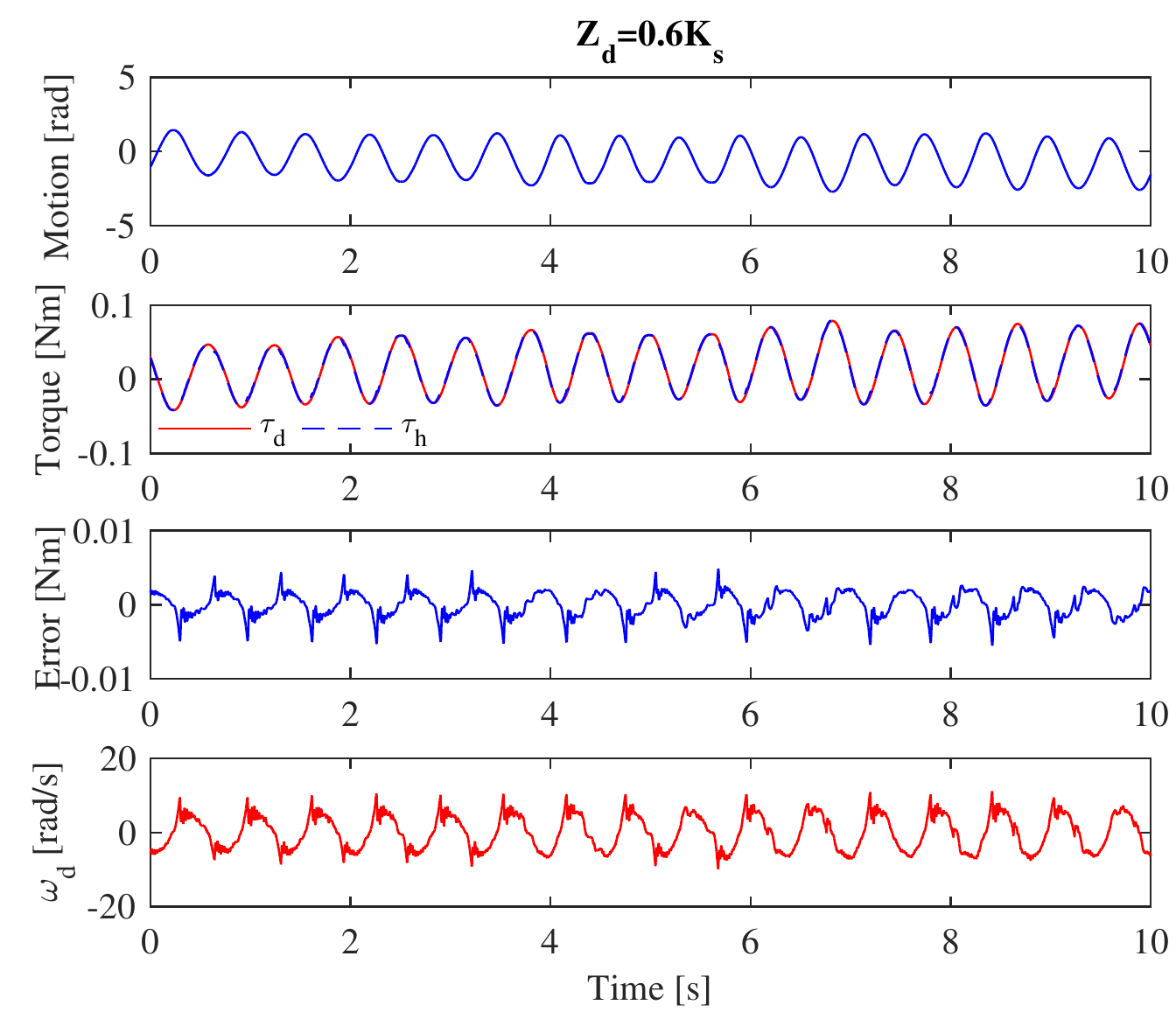}}\quad\quad\quad
	\subfigure[]{\includegraphics[width=0.82\columnwidth]{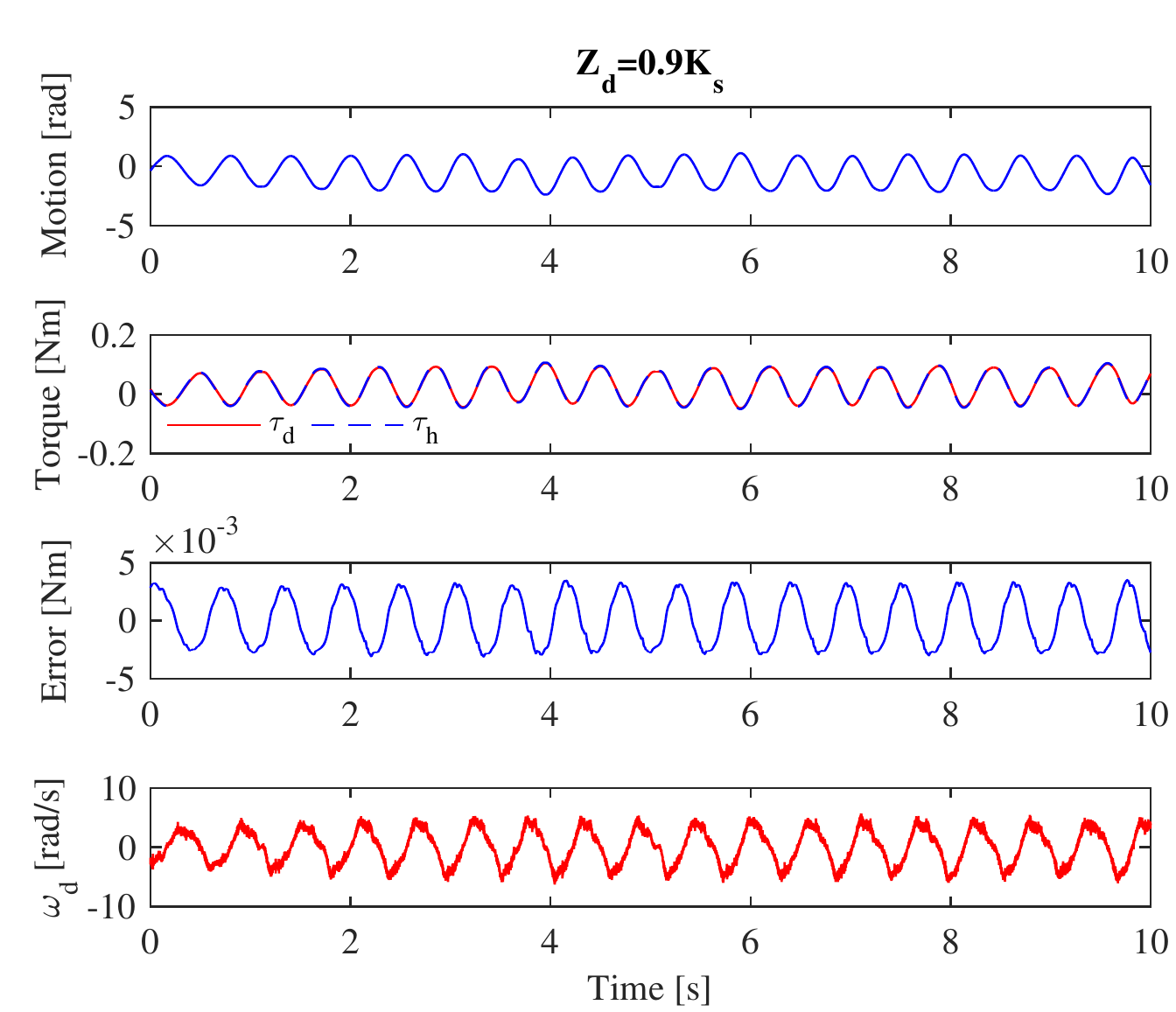}}
	\caption{Experimental results of the $H_\infty$ synthesis-based method (3rd-order controller) for different stiffness control.}\label{fig_Experiment}
\end{figure*}

The stiffness rendering results including the hand motion $\varphi_h$, the actual interaction torque $\tau_h$, the desired torque $\tau_d$, the torque error $e$, the desired motor velocity $\omega_d$ were all drawn in Fig.~\ref{fig_Experiment} for the desired stiffness $0$, $0.3K_s$, $0.6K_s$ and $0.9K_s$. The respective bounds $\gamma_1$ were set as 0.054, 0.029, 0.016 and 0.004. The actual interaction torque $\tau_h$ tracked the desired torque $\tau_d$ well with small torque error, indicating good stiffness rendering performance. 

As summarized in Table~\ref{table2}, the respective maximal torque errors were 0.0149 Nm, 0.0116 Nm, 0.0055 Nm, 0.0028 Nm, the respective sums of squared error (SSE) were 1.4238 (Nm)$^2$, 0.8262 (Nm)$^2$, 0.0561 (Nm)$^2$, 0.0341 (Nm)$^2$, the respective maximal desired velocities were 19.5 rad/s, 15.6 rad/s, 11.2 rad/s, 5.8 rad/s, the respective signal to noise ratios (SNR) of the desired velocity were 18.3 dB, 15.8 dB, 13.9 dB, 12.3 dB. The maximal torque error, SSE, the maximal desired velocity and SNR decreased as the desired stiffness $Z_d$ increased. The desired velocities were all within the motor saturation limit, and had high signal to noise ratios.

\begin{table*}[!htbp]
	\centering
	\caption{Comparison of quantified metrics between the passivity-based PID and $H_\infty$ synthesis methods for different stiffness control during experiments. PID: passivity-based PID method, $H_\infty$-PID: $H_\infty$ synthesis-based PID, $H_\infty$: $H_\infty$ synthesis with 3rd-order controller. SSE: sum of squared error, SNR: signal to noise ratio.}
	\footnotesize
	\tabcolsep=5.5pt
	\label{table2}
	\begin{center}
		\renewcommand{\arraystretch}{1.5}
		\begin{tabularx}{0.8\textwidth}{c|cc|cc|ccc|cc}
			\hline
			& \multicolumn{2}{c|}{0}  & \multicolumn{2}{c|}{$0.3K_s$} & \multicolumn{3}{c|}{$0.6K_s$} & \multicolumn{2}{c}{$0.9K_s$}  \\
			& $H_\infty$ & PID & $H_\infty$ & PID & $H_\infty$ & $H_\infty$-PID & PID & $H_\infty$ & PID\\
			\hline
			Maximal $e$ (Nm)   &0.0149 &  0.0221 & 0.0116 & 0.0143 & 0.0055 & 0.0065 & 0.0095 & 0.0028 & 0.0057\\
			SSE ((Nm)$^2$) & 1.4238 & 3.1702 & 0.8262 & 1.5970 & 0.0561 & 0.1677 & 0.5241 & 0.0341 & 0.0411\\
			Maximal $\omega_d$ (rad/s) &19.5 & 24.6 & 15.6 & 19.4 & 11.2 & 11.3 & 14.9 & 5.8 & 12.9  \\
			SNR of $\omega_d$ (dB) & 18.3 & 16.2 & 15.8 & 13.2 & 13.9 & 12.1 & 10.2 & 12.3 & -4.5\\
			\hline
		\end{tabularx}
	\end{center}
\end{table*}

To accurately render a smaller stiffness, such as zero impedance, higher control effort and quicker velocity response from the motor were demanded for the cable-driven SEA to respond to the human motion. But for the practical system, motor performance limitations may be the major restrictions for the rendering accuracy. When rendering a stiffness that closes to the physical stiffness of the elastic component, smaller torque error and less actuation effort were achieved due to the smaller error between the desired and the physical stiffness.

The system had backlash when the handle motion crosses the zero point. It was mainly caused by the mechanical gap between the motor output shaft and the cable winch. The backlash could cause short oscillations in the control signal  when crossing the zero point as shown in Fig.~\ref{fig_Experiment} and \ref{fig_Experiment_Comparison}. In our control strategy, the backlash can be treated as noise signal since it can be detected by the position encoder. To obtain good noise rejection, we can put the two frequency-domain constraints (\ref{eq_req_n}) and (\ref{eq_req_nu}) to the system. In our experiments, the interaction torque remained smooth when crossing the zero point.

\begin{figure}[!htbp]
	\centering
	\includegraphics[width=0.9\columnwidth]{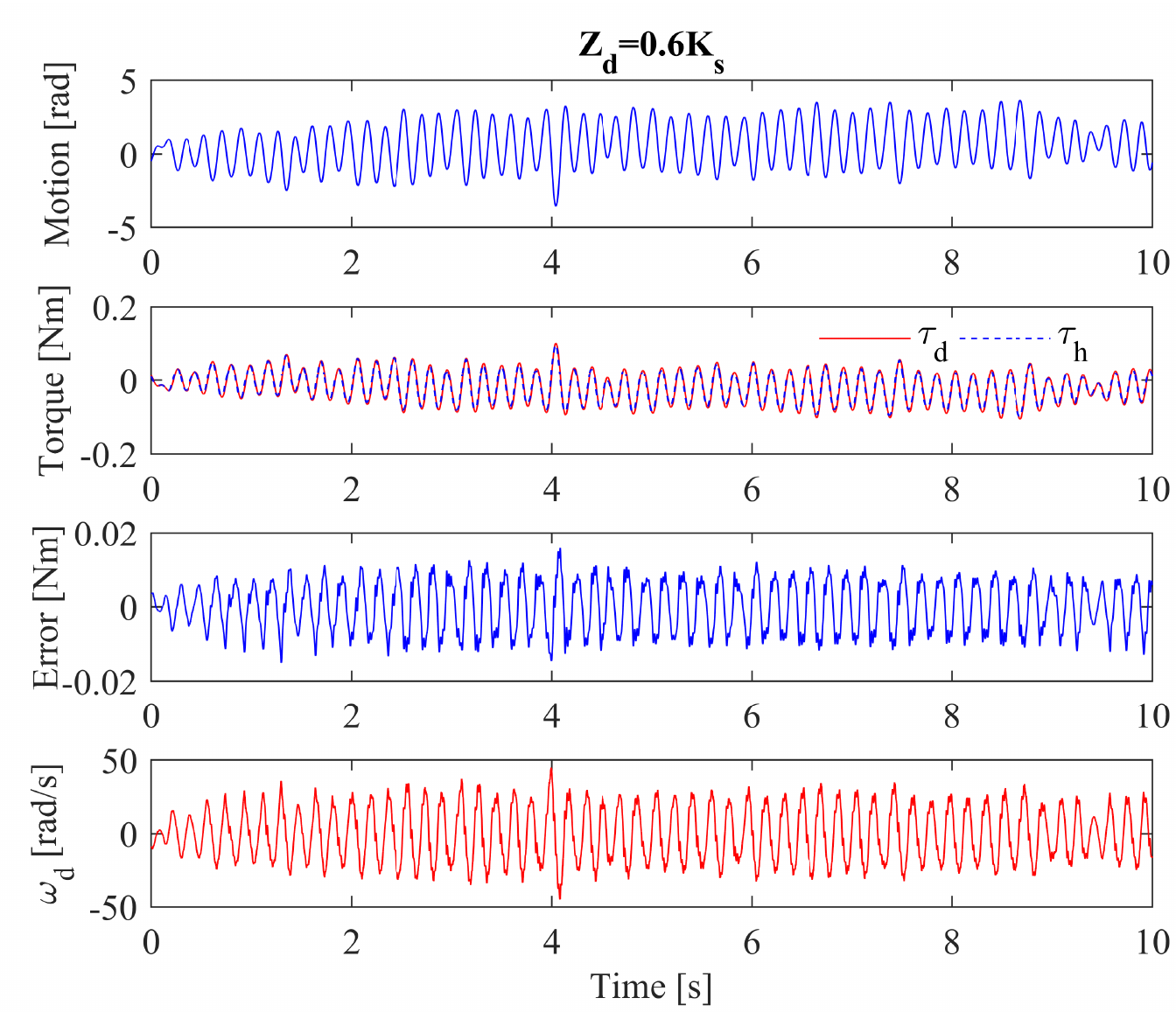}
	\caption{Experimental results of $H_\infty$ synthesis-based stiffness control (3rd-order controller) for $Z_d=0.6K_s$ when the motion $\varphi_h$ varying up to 6 Hz.}\label{fig_Experiment_High_Freq}
\end{figure}

To demonstrate the stiffness control performance at a higher frequency range, experimental results for $Z_d=0.6K_s$ were presented in Fig.~\ref{fig_Experiment_High_Freq}. The human hand motion frequency varied up to about 6 Hz. The quantified metrics including maximal torque error, SSE, maximal desired velocity and SNR were 0.0158 Nm, 0.8901 (Nm)$^2$, 44.9 rad/s and 13.4 dB, respectively.

\subsection{Comparison with the Passivity-based PID  Method}

To further illustrate the benefit of the $H_\infty$ synthesis-based stiffness control method, a passivity-based PID method and an $H_\infty$ synthesis-based PID method were used for comparison. 

Detailed analysis of the passivity-based PID method can be found in~\cite{Vallery2008,Sergi2015,Tagliamonte2014,Calanca2017}. To guarantee the passivity conditions as shown in Section \ref{Section_Passivity} and to obtain good stiffness control performance, the PID controller was tuned as
\begin{equation}
\label{eq_PID}
K_{PID}(s)=-(1000+\frac{10}{s}+20s).
\end{equation}

For the $H_\infty$ synthesis-based PID method, the derivative term in the PID controller was replaced by a filtered derivative. The frequency-domain constraints were the same as the $H_\infty$ synthesis method with 3rd-order controller when $Z_d=0.6K_s$.  With the $H_\infty$ synthesis, the PID controller was given as
\begin{equation}
\label{eq_H_infty_PID}
K_{HPID}(s)=-(1333+\frac{0.0002}{s}+\frac{403s}{s+8}).
\end{equation}

The maximal torque errors, SSEs, maximal desired velocities, SNRs for different stiffness control during experiments were also summarized in Table~\ref{table2}.

Comparison between the passivity-based PID method and the $H_\infty$ synthesis-based method with 3-rd order controller showed that the proposed $H_\infty$ method achieved smaller maximal error, SSE, maximal desired velocity and larger SNR for each case, indicating more accurate and robust stiffness control with less control effort. Especially for the case of $Z_d=0.9K_s$, the SNR of the passivity-based PID method was $-4.5$ dB below zero, indicating that the noise in the desired velocity almost overwhelmed the signal.

\begin{figure}[!htbp]
	\centering
	\subfigure[]{\includegraphics[width=0.85\columnwidth]{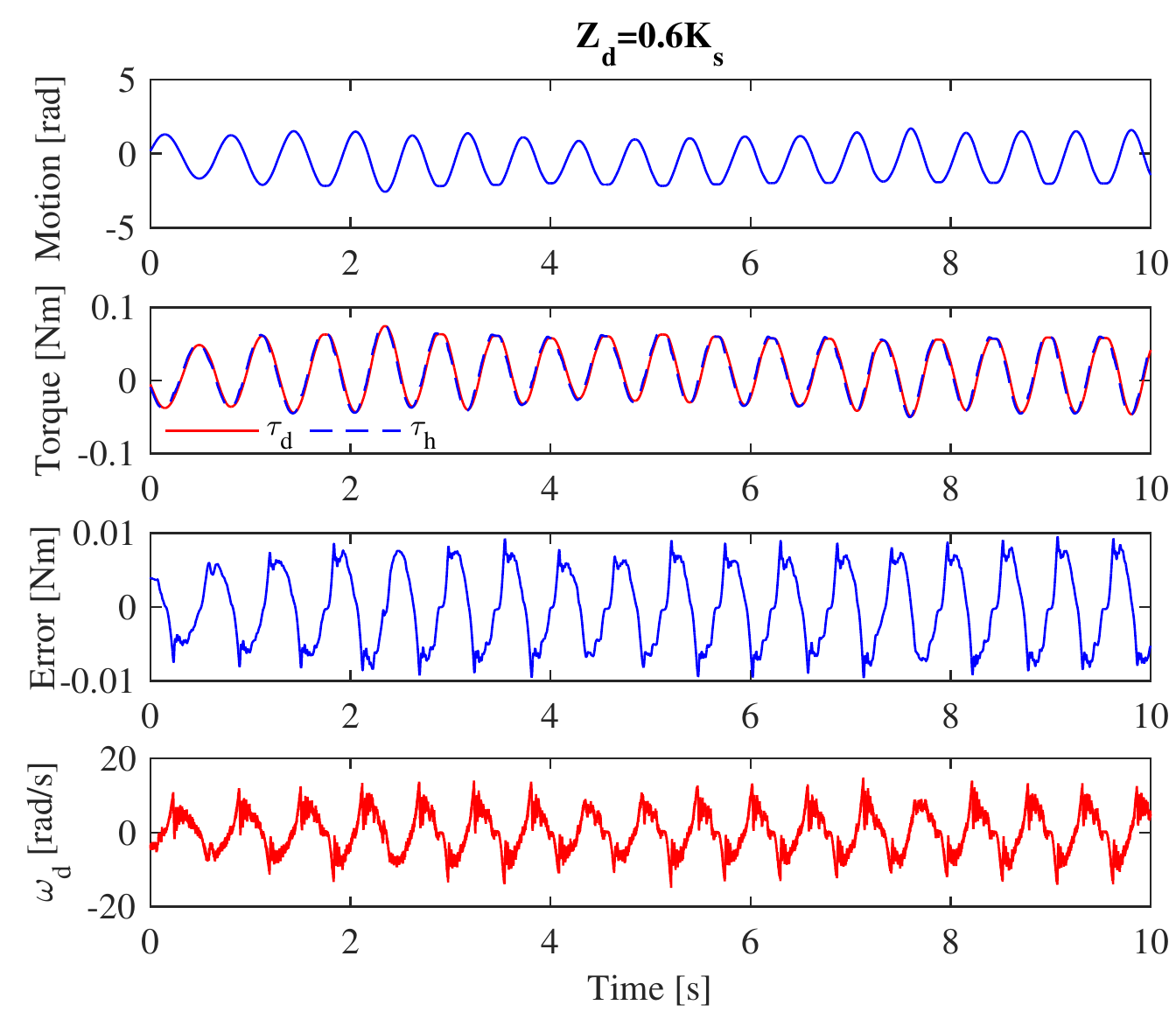}}\\
	\subfigure[]{\includegraphics[width=0.85\columnwidth]{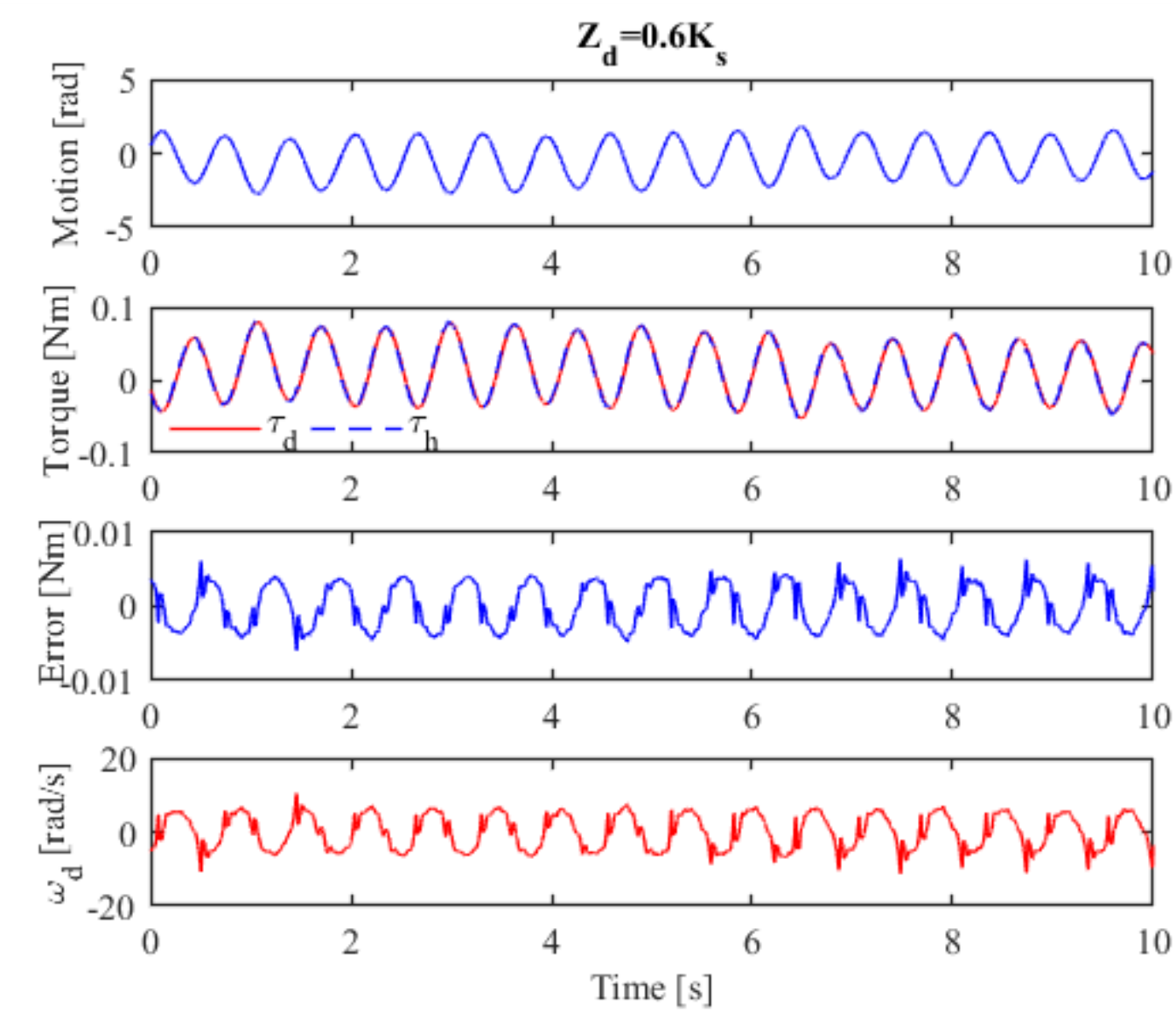}}\\
	\subfigure[]{\includegraphics[width=0.85\columnwidth]{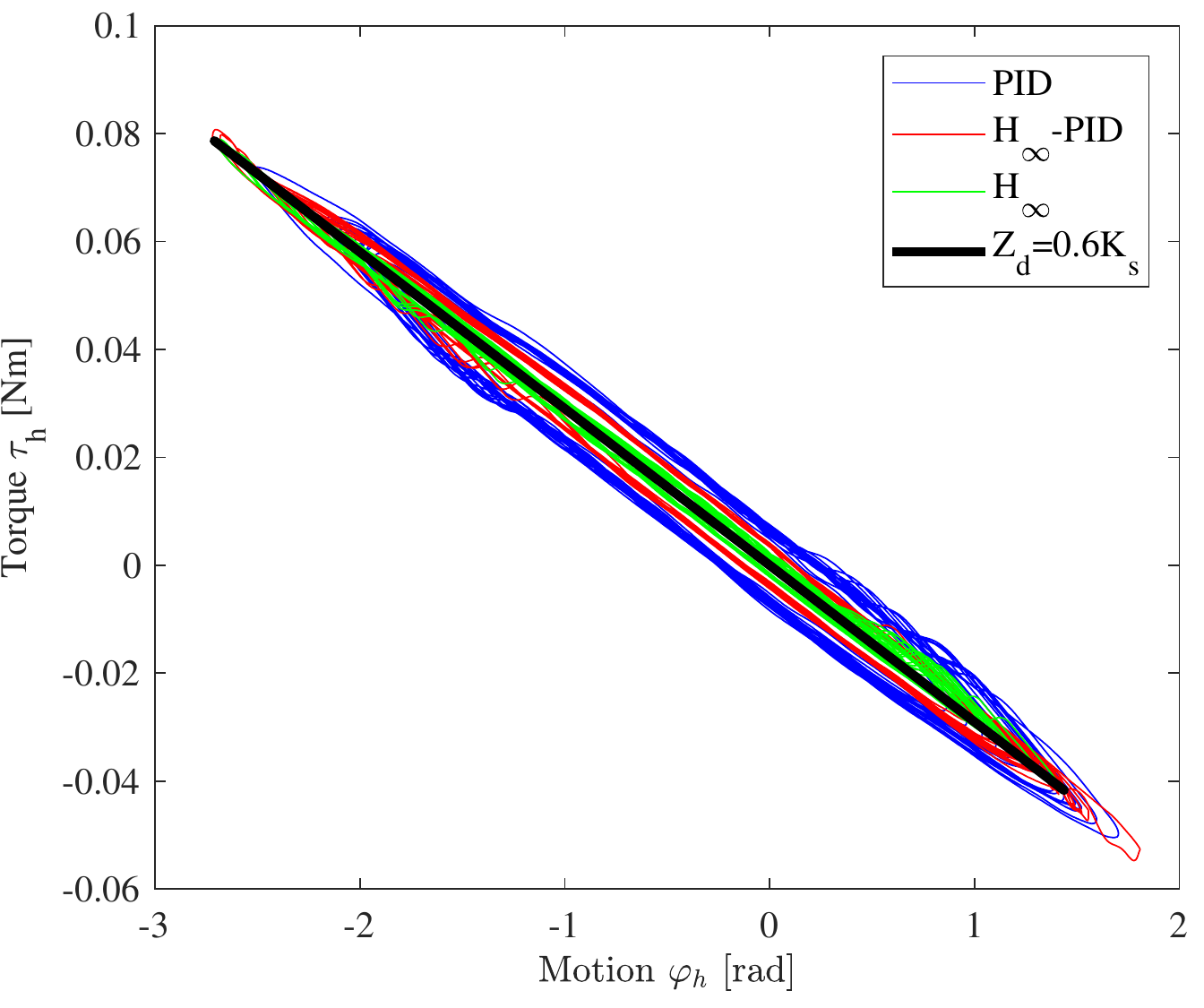}}	
	\caption{(a) Experimental results of passivity-based PID method for $Z_d=0.6K_s$. (b) Experimental results of $H_\infty$ synthesis-based PID method for $Z_d=0.6K_s$. (c) Comparison of stiffness accuracy between the three methods for $Z_d=0.6K_s$. PID: passivity-based PID method, $H_\infty$-PID: $H_\infty$ synthesis-based PID, $H_\infty$: $H_\infty$ synthesis with 3rd-order controller.}
	\label{fig_Experiment_Comparison}
\end{figure}

When $Z_d=0.6K_s$, the stiffness control results of the passivity-based and $H_\infty$ synthesis-based PID method were depicted in Fig.~\ref{fig_Experiment_Comparison}(a) and (b). Comparison of Fig.~\ref{fig_Experiment_Comparison}(a) with Fig.~\ref{fig_Experiment_Comparison}(b) and Fig.~\ref{fig_Experiment}(c) showed the controller output $\omega_d$ of the passivity-based PID method was much more noisy than our $H_\infty$ method because the pure derivative term in the PID controller can amplify the sensor noise. However, to ensure passivity, the derivative term should be big enough.

Further observation from Fig.~\ref{fig_Experiment_Comparison}(c) and Table~\ref{table2}, the overall deviation from the desired stiffness was much more smaller by the $H_\infty$ method. The $H_\infty$ synthesis-based PID method was much more accurate and robust than the passivity-based PID method, but less accurate than the $H_\infty$ synthesis method with 3-rd order controller. Actually, the PID controller with filtered derivative can be viewed as a 2nd-order controller, which was a special case when $n_k=2$ for the $H_\infty$ synthesis method.

\section{Conclusion and Future Work}
\label{section5}

In this work, we have enhanced the rendering performance of a cable-driven SEA with restricted frequency-domain specifications for physical human-robot interaction. The stiffness control problem was reformulated into an $H_\infty$ controller synthesis framework. A structured dynamic output-feedback controller was synthesized to achieve accurate impedance rendering performance in the specified low-frequency range in which the human interactive motion is dominant, to reject noise at the high-frequency domain, to attenuate disturbance at the full-frequency band, and especially to guarantee passive interaction with external dynamic motion. It achieved good stiffness rendering performance for various desired stiffnesses both in simulations and experiments. Compared with the passivity-based PID method, the proposed method obtained more accurate and robust stiffness rendering.

It will be difficult to design a well-performed controller satisfying a large number of requirements. If the constraints cannot be fulfilled at the same time, a tuning process is needed and this process can be heavy. Besides, variable stiffness control where the desired stiffness is time-varying, brings challenges on stability and passivity. The future work will be focused on these aspects.

\section*{Acknowledgments}

The authors thank Prof. Li Qiu from Hong Kong University of Science and Technology and Prof. Xiang Chen from University of Windsor for the helpful suggestions.

This work was supported by the National Natural Science Foundation of China (61720106012), the Foundation of State Key Laboratory of Robotics (2016-003) and the Fundamental Research Funds for the Central Universities.

\bibliography{reference}

\end{document}